\documentclass[preprint,12pt]{elsarticle}
\usepackage{geometry}
\usepackage{graphicx}
\usepackage{amsmath}
\usepackage{amssymb}
\usepackage{enumitem}
\usepackage{float}
\usepackage{xcolor}

\usepackage[hidelinks]{hyperref}
\usepackage[pagewise]{lineno}

\usepackage{multirow}
\usepackage{caption}
\usepackage{adjustbox}
\usepackage{tabularx}
\usepackage{subcaption}
\usepackage{longtable}
\usepackage{setspace}
\usepackage{tabularray}
\usepackage{makecell}

\journal{Travel Behaviour and Society}
\begin{document}

\newcommand{\YX}{\color{black}} 
\newcommand{\YXRII}{\color{black}} 

\begin{frontmatter}



\title{Benchmarking Retrieval-Augmented Generation Strategies for Large Language Model-Based Travel Mode Choice Prediction}

\author[UT]{Yiming Xu\corref{cor1}}
\author[UT]{Junfeng Jiao}

\affiliation[UT]{organization={School of Architecture, The University of Texas at Austin},
            addressline={310 Inner Campus Drive, B7500}, 
            city={Austin},
            postcode={78712}, 
            state={TX},
            country={United States}}

\cortext[cor1]{Corresponding author. Address: 310 Inner Campus Drive, B7500, Austin, TX 78712. Phone: +01 352-871-3481. Email: yiming.xu@utexas.edu}

\begin{abstract}
Accurately predicting travel mode choice is essential for effective transportation planning, yet traditional statistical and machine learning models are constrained by rigid assumptions, limited contextual reasoning, and reduced {\YX transferability}. This study explores the potential of Large Language Models (LLMs) as a more flexible and context-aware approach to travel mode choice prediction, enhanced by Retrieval-Augmented Generation (RAG) to ground predictions in empirical data. We develop a modular framework for integrating RAG into LLM-based travel mode choice prediction and evaluate four retrieval strategies: basic RAG, RAG with balanced retrieval, RAG with a cross-encoder for re-ranking, and RAG with balanced retrieval and cross-encoder for re-ranking. These strategies are tested across three LLM architectures (OpenAI GPT-4o, o4-mini, and o3) to examine the interaction between model reasoning capabilities and retrieval methods. Using the 2023 Puget Sound Regional Household Travel Survey data, we conduct a series of experiments to evaluate model performance. The results demonstrate that RAG substantially enhances predictive accuracy across a range of models. Notably, the GPT-4o model combined with balanced retrieval and cross-encoder re-ranking achieves the highest accuracy of 80.8\%, exceeding that of conventional statistical and machine learning baselines. Furthermore, LLM–based models exhibit superior {\YX zero-shot transfer abilities} relative to these baselines. Findings highlight the critical interplay between LLM reasoning capabilities and retrieval strategies, demonstrating the importance of aligning retrieval strategies with model capabilities to maximize the potential of LLM-based travel behavior modeling. 

\end{abstract}




\end{frontmatter}


\section{Introduction}
\label{s:1}
The process by which individuals select a particular mode of transportation for a trip, known as travel mode choice, stands as one of the most critical and extensively studied components within the broader field of travel demand forecasting \citep{ben1985discrete}. The accurate prediction of these choices is not only a technical exercise in transportation modeling but also a fundamental prerequisite for effective urban planning, sustainable development, and the enhancement of public well-being \citep{handy1996methodologies, schneider2013theory}. The decisions individuals make—whether to drive, take public transit, cycle, or walk—aggregate into system-level outcomes that significantly shape the urban fabric. These collective choices dictate the demand for transportation infrastructure, influence the severity of traffic congestion, determine the environmental footprint of mobility, and ultimately define the quality of life within a city. Consequently, the ability to develop robust and reliable travel mode choice prediction models is essential for policymakers and planners striving to create efficient, equitable, and livable urban environments. 

Predicting travel mode choice is inherently complex, influenced by traveler demographics, trip characteristics, and transportation options \citep{barff1982selective, paulssen2014values, buehler2011determinants, bhat2000incorporating}. Traditionally, discrete choice models like the Multinomial Logit (MNL), grounded in random utility maximization, have provided interpretable behavioral insights through model coefficients \citep{ben1985discrete, ben1999discrete, bhat1995heteroscedastic}. However, their rigid assumptions often fail to capture the non-linear dynamics of human behavior, leading to biased predictions. Machine learning models, such as Random Forests, improve predictive accuracy by learning complex patterns directly from data but at the cost of interpretability \citep{hagenauer2017comparative, zhao2020prediction, kashifi2022predicting}. Moreover, both approaches rely on structured numerical data and lack contextual reasoning, limiting their {\YX transferability} to novel situations or policies beyond their training scope.

The recent emergence of Large Language Models (LLMs) presents a new paradigm with the potential to resolve these limitations. Pre-trained on vast corpora of text, LLMs possess a unique capacity for context-awareness, allowing them to interpret unstructured, qualitative information \citep{achiam2023gpt}. More importantly, LLMs exhibit emergent reasoning capabilities. Through techniques like Chain-of-Thought (COT), they can articulate a step-by-step rationale for their predictions in natural language, offering a transparent and intuitive form of explainability \citep{xu2025largereasoningmodelssurvey}. Furthermore, their strong zero-shot and few-shot learning abilities make them highly data-efficient and generalizable.

Despite their power, LLMs are prone to generating hallucinations, and their knowledge is limited to their training data. To overcome this for high-stakes applications like transportation planning, Retrieval-Augmented Generation (RAG) has emerged as a critical enabling technology. RAG enhances an LLM by dynamically retrieving relevant, external information—in this case, similar past trips from a database—and providing it to the model as context for its prediction \citep{lewis2020retrieval}. This process grounds the LLM's reasoning in empirical data, mitigating the risk of hallucination and transforming a general-purpose model into a specialized, data-driven decision-making tool \citep{gao2024retrieval}.

While the potential of LLMs and RAG is clear, their application to travel mode choice prediction remains limited. This study addresses two critical research gaps. First, there is a need for a systematic evaluation of advanced RAG strategies to understand how different retrieval methods, from basic retrieval to more sophisticated pipelines, impact predictive accuracy. Second, the interplay between the choice of the underlying LLM architecture and the RAG strategy has not been examined. It is unclear whether a more powerful base model is always superior, or if a less advanced model can achieve better performance when paired with a highly optimized retrieval system. Answering this is essential for establishing best practices for deploying these models effectively.


{\YX
More importantly, this study seeks to move beyond a standard tabular-to-text and in-context learning pipeline by examining how retrieval should be designed for travel behavior modeling. Travel survey datasets are commonly characterized by strong class imbalance, where dominant modes such as driving and walking are substantially more frequent than transit, cycling, or micromobility. Under such conditions, naive similarity-based retrieval may reinforce majority-class bias by retrieving examples primarily from dominant modes. In addition, the benefit of retrieval augmentation may depend on the reasoning capability of the underlying LLM. While retrieved examples can ground models in empirical data, they may also introduce noise for high-reasoning models if the retrieved context is biased or weakly relevant. This creates an important need to understand when retrieval helps or hurts, and how retrieval quality can be improved through strategies such as balanced retrieval and cross-encoder re-ranking.

This paper aims to fill these gaps through a comprehensive benchmark analysis of RAG strategies for LLM-based travel mode choice prediction. The methodological contribution of this study lies not only in applying LLMs to travel mode choice prediction, but also in deriving practical retrieval-design insights for travel behavior applications. 
Specifically, the key contributions are threefold. First, we develop a modular and generalizable framework that integrates RAG with LLMs for travel mode choice prediction, systematically evaluating zero-shot prediction, basic RAG, RAG with balanced retrieval, RAG with cross-encoder re-ranking, and RAG with balanced retrieval combined with cross-encoder re-ranking. Second, we provide a class-imbalance-aware evaluation of retrieval design by examining whether balanced retrieval improves the prediction of underrepresented travel modes. Third, we investigate the interaction between retrieval strategy and LLM reasoning capability across GPT-4o, o4-mini, and o3, showing that retrieval augmentation is not universally beneficial and that high-reasoning models require more precise retrieved evidence to outperform their zero-shot baselines. Together, these analyses provide methodological guidance for deploying RAG-augmented LLMs in travel behavior modeling, including how to balance accuracy, minority-class performance, runtime, token usage, and cost.

}

\section{Literature Review}
\label{s:2}

\subsection{Travel Mode Choice Modeling}

The study of travel mode choice is a mature field in transportation research, recognized as a critical step in travel demand forecasting \citep{ben1985discrete, ben1999discrete, hasnine2021tour}. The choice of a travel mode is a complex decision influenced by a wide array of factors. These are broadly categorized into three groups: characteristics of the traveler, features of the trip, and attributes of the transportation system and built environment \citep{buehler2011determinants, bhat2000incorporating, eldeeb2021built}. Traveler characteristics include socio-demographic and economic factors like age, gender, income, and vehicle ownership. Trip characteristics encompass variables such as travel distance, time of day, and trip purpose (e.g., work, school, leisure). Finally, system and environmental attributes include the level of service of different modes (travel time, cost), the quality of infrastructure (e.g., availability of sidewalks and bike lanes), and exogenous factors like weather conditions and land use patterns.

Based on these factors, two primary paradigms have been developed for travel mode choice modeling: statistical models and machine learning models. For decades, the field has relied on discrete choice models, particularly the Multinomial Logit and Nested Logit models \citep{ben1985discrete, ben1999discrete, bhat1995heteroscedastic, bhat2000incorporating}. Rooted in the economic theory of random utility maximization, the primary strength of logit models is their high degree of interpretability. Their coefficients have clear behavioral meanings, allowing planners to understand why choices are made. However, logit models are built on strict statistical assumptions, such as the Independence of Irrelevant Alternatives (IIA) property, and they struggle to capture the complex, non-linear relationships inherent in human behavior, which can lead to biased predictions \citep{zhao2020prediction}.

In response to the predictive shortcomings of statistical models, researchers have increasingly turned to machine learning techniques such as random forests, gradient boosting methods, and neural networks \citep{hagenauer2017comparative, zhao2020prediction, lee2018comparison, chen2023travel, cheng2019applying}. The primary advantage of these models is their superior predictive accuracy, as their data-driven nature allows them to learn complex, non-linear patterns. However, this predictive power is typically achieved at the expense of interpretability, leading to the well-known "black box" problem \citep{zhao2020prediction, xu2021identifying}. While an ML model may accurately predict what mode an individual will choose, its internal logic is often opaque, limiting its utility for policy analysis.

In addition, both statistical and machine learning models are limited by their very nature as mathematical constructs operating on structured, numerical data. They lack any genuine understanding of the real-world context of a trip or the lived experience of the traveler \citep{ge2025llm}. This absence of contextual reasoning limits the {\YX transferability} of these models, especially when attempting to predict behavior in novel situations or under policy regimes that fall outside the distribution of the training data.

\subsection{LLMs in Travel Behavior Modeling}

The recent emergence of large language models has introduced a new frontier for modeling complex human behaviors, including discrete choice and decision-making \citep{ge2025llm, jiawei2024large, gong2024mobility, liu2025toward, adornetto2025generative, liu2025trip, ying2025beyond}. Unlike traditional models that require structured numerical inputs, LLMs can process and reason over rich, descriptive text. This capability allows them to potentially capture more nuanced aspects of a choice context. Studies have shown that LLMs can achieve impressive performance on various decision-making tasks by leveraging in-context learning, where examples are provided in the prompt to guide the model's prediction \citep{ge2025llm, zhang2025transmode}. This approach, often referred to as zero-shot or few-shot prompting, has been explored for tasks ranging from consumer product choice to policy decisions, demonstrating the potential of LLMs as "zero-data" classifiers.

The application of LLMs specifically to travel mode choice prediction is a nascent but promising area of research. Initial studies suggest that LLMs can predict travel modes by reasoning over textual descriptions of trip scenarios without requiring explicit model training \citep{zhang2025transmode, nie2025exploring, liu2024can}. This presents a significant departure from the data-intensive and statistically rigid methods that have historically dominated the field. However, the "zero-shot" performance of LLMs in this domain is often constrained by the model's internal knowledge, which may not perfectly align with the specific geographic or demographic context of a given travel survey \citep{liu2024can}. The models can be sensitive to prompt phrasing and may not consistently outperform finely-tuned traditional models without specialized adaptation. To address these shortcomings, researchers are exploring retrieval-augmented strategies. Retrieval-Augmented Generation (RAG) enhances LLM predictions by providing the model with relevant examples retrieved from a specific dataset, effectively grounding the model's reasoning in contextually appropriate information \citep{lewis2020retrieval}. However, few existing studies have explored the use of RAG to enhance LLMs' predictions for travel mode choice. This study builds upon this emerging body of work by systematically evaluating how different RAG strategies and LLM architectures can be optimized to improve the accuracy of travel mode choice prediction, moving beyond simple zero-shot approaches to create more robust and context-aware models.

\section{Methodology}
\label{s:3}
\subsection{Problem Formulation}
In this study, the task of travel mode choice prediction is formally defined as a multi-class classification problem. The objective is to predict the transportation mode that an individual traveler will choose for a specific trip, given a set of explanatory variables describing both the traveler and trip characteristics. It is worth noting that, unlike the conventional discrete choice modeling framework, which necessitates information on all available alternatives and their attributes, this formulation relies solely on the information associated with the observed mode. This multi-class classification formulation enables the use of broader datasets and consistent with existing data-driven studies using large-scale travel survey data \citep{kashifi2022predicting, xia2023random, hagenauer2017comparative, cheng2019applying, chapleau2019application, wang2018machine}.

Consider a traveler $n$ from a set of $N$ individuals who is making a trip $t$. The traveler faces a discrete choice set $M$ containing $k$ independent travel mode alternatives, such that $M=\{m_1,m_2,\dots,m_k\}$. We assume that all travelers share the same set of choices $M$. The choice made by traveler $n$ for trip $t$ is denoted by $y_{nt}$, where $y_{nt}\in M$. The decision-making process is influenced by a vector of input features, $X_{nt}$, which represents the context of the choice. This feature vector is composed of several categories of attributes, including traveler characteristics (e.g., age, gender, income) and trip characteristics (e.g., distance, purpose). 

The core problem is to learn a mapping function, $f$, that takes the feature vector $X_{nt}$ as input and outputs a predicted travel mode, $\hat{y}_{nt}$, such that
\begin{align}
\hat{y}_{nt} &= f(X_{nt})
\end{align}

In a probabilistic framework, the model estimates the conditional probability $P(y_{nt} = m | X_{nt})$ for each mode $m \in M$. The final predicted mode is the one that maximizes this probability:
\begin{align}
\hat{y}_{nt} &= \underset{m \in M}{\operatorname*{argmax}} \; P(y_{nt} = m \mid X_{nt})
\end{align}

This general formulation applies to traditional discrete choice models, conventional machine learning classifiers, and the LLM-based approaches investigated in this study. In the context of LLMs, the structured feature vector $X_{nt}$ is transformed into a natural language prompt that describes the choice scenario. The LLM then processes this textual input to generate the prediction $y_{nt}$, leveraging its reasoning and context-awareness capabilities to perform the classification task. 

\subsection{Travel Mode Choice Prediction Framework}

\begin{figure}[!ht]
\centering
\includegraphics[width=\textwidth]{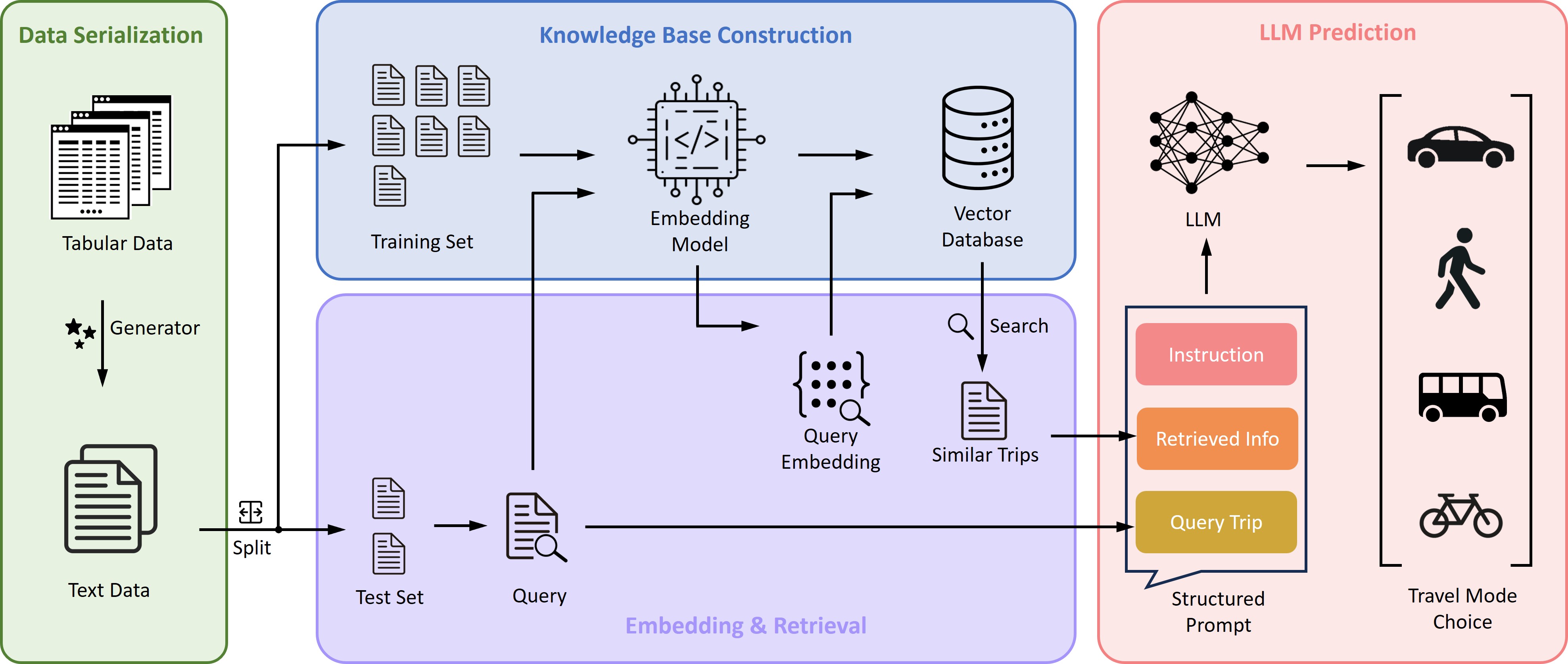} 
\caption{Framework for travel mode choice prediction using RAG augmented LLM}
\label{fig:framework}
\end{figure}

To systematically evaluate the performance of LLMs in travel mode choice prediction, we developed a modular framework with RAG. This framework is designed to ingest structured tabular data, construct a searchable knowledge base of past travel behaviors, and employ various retrieval strategies to elicit reasoned predictions from an LLM. As presented in Figure~\ref{fig:framework}, the architecture comprises a multi-stage pipeline, starting with a data serialization phase and culminating in a retrieval and generation phase to make predictions for new, unseen trips.

The process begins with data serialization, which converts tabular survey data into a text format that the model can natively process. Each record in the tabular dataset, representing a unique trip with its associated traveler characteristics, is converted into a coherent, human-readable natural language description. This table-to-text generation process uses a template-based approach where feature names and their corresponding values are woven into sentences (e.g., "The trip distance is 5 miles. The traveler is 35 years old."). Once serialized, the entire dataset is split into a training set and a test set to ensure a rigorous evaluation of the model's ability to generalize. Note that the travel mode choice for trips in the test set is masked to avoid information leakage. 

After that, the serialized training set is used to construct the external knowledge base. Each text description of a trip in the training data is treated as a document. These documents are then processed by an embedding model, which converts the text into high-dimensional vector representations. These vectors are subsequently indexed and stored in a specialized vector database. For this framework, we utilize Facebook AI Similarity Search (FAISS), a library optimized for highly efficient similarity search of dense vectors \citep{douze2025faisslibrary}. This indexed vector store serves as the retriever, providing a searchable knowledge base of past trips that the LLM can query at inference time.

At the core of the framework is a retrieval-and-generation process executed for each trip in the test set. The retrieval stage embeds the serialized text of a test trip (the query) and performs a similarity search against the FAISS index to identify the most analogous trips in the vector store. The key goal of this study is to compare the performance of different retrieval strategies within this framework. Accordingly, we evaluate four distinct RAG strategies. The first is a basic RAG that retrieves the top-most similar examples. The second incorporates balanced retrieval, which addresses class imbalance by selecting a more diverse set of examples across travel modes. The third employs a cross-encoder re-ranking mechanism, implementing a two-stage process that refines the initial retrieval results to maximize relevance. Finally, the fourth strategy combines balanced retrieval with cross-encoder re-ranking. Detailed descriptions of these approaches are provided in the following section.

Finally, the documents obtained from the selected retrieval strategy are used to construct an augmented prompt for the LLM. This prompt is carefully engineered with three key components: an instruction that clearly defines the travel mode choice prediction task, the context containing the retrieved, similar trip descriptions, and the input data representing the test trip itself. {\YX The prompt template is:

\begin{quote}
\textbf{System message:}\\
\textit{You are a travel behavior expert. Your task is to predict the most likely travel mode for a trip.}

\vspace{0.5em}
\textbf{User message:}\\
\textit{Trip details:}\\
\{serialized trip description\}

\vspace{0.5em}
\textit{Relevant past trips:}\\
\{retrieved trip 1, including its observed trip mode\}\\
\{retrieved trip 2, including its observed trip mode\}\\
...\\
\{retrieved trip K, including its observed trip mode\}

\vspace{0.5em}
\textit{Question: Based on the trip details and the relevant past trips, what is the most likely trip mode?\\
Only output one of: [Drive, Walk, Transit, Bike/Micromobility].}
\end{quote}

The LLM response was stripped of leading and trailing whitespace and normalized by converting it to lowercase and removing extraneous punctuation. If the response exactly matched one of the four valid labels, it was directly mapped to the corresponding class. If the response contained a valid label together with additional text, the valid label was extracted. Outputs referring to common synonyms were mapped to the canonical labels, such as “Driving” to “Drive”, “Bus” to “Transit”, and “Bicycle” to “Bike/Micromobility”. If no valid label could be identified, the prediction was treated as invalid.

}

By providing the LLM with these relevant, in-context examples, the model is grounded in empirical data from the knowledge base. This process enhances the accuracy and contextual awareness of the LLM's final prediction while mitigating the risk of hallucination.

\subsection{RAG Strategies}
\begin{figure}[!ht]
\centering
\includegraphics[width=\textwidth]{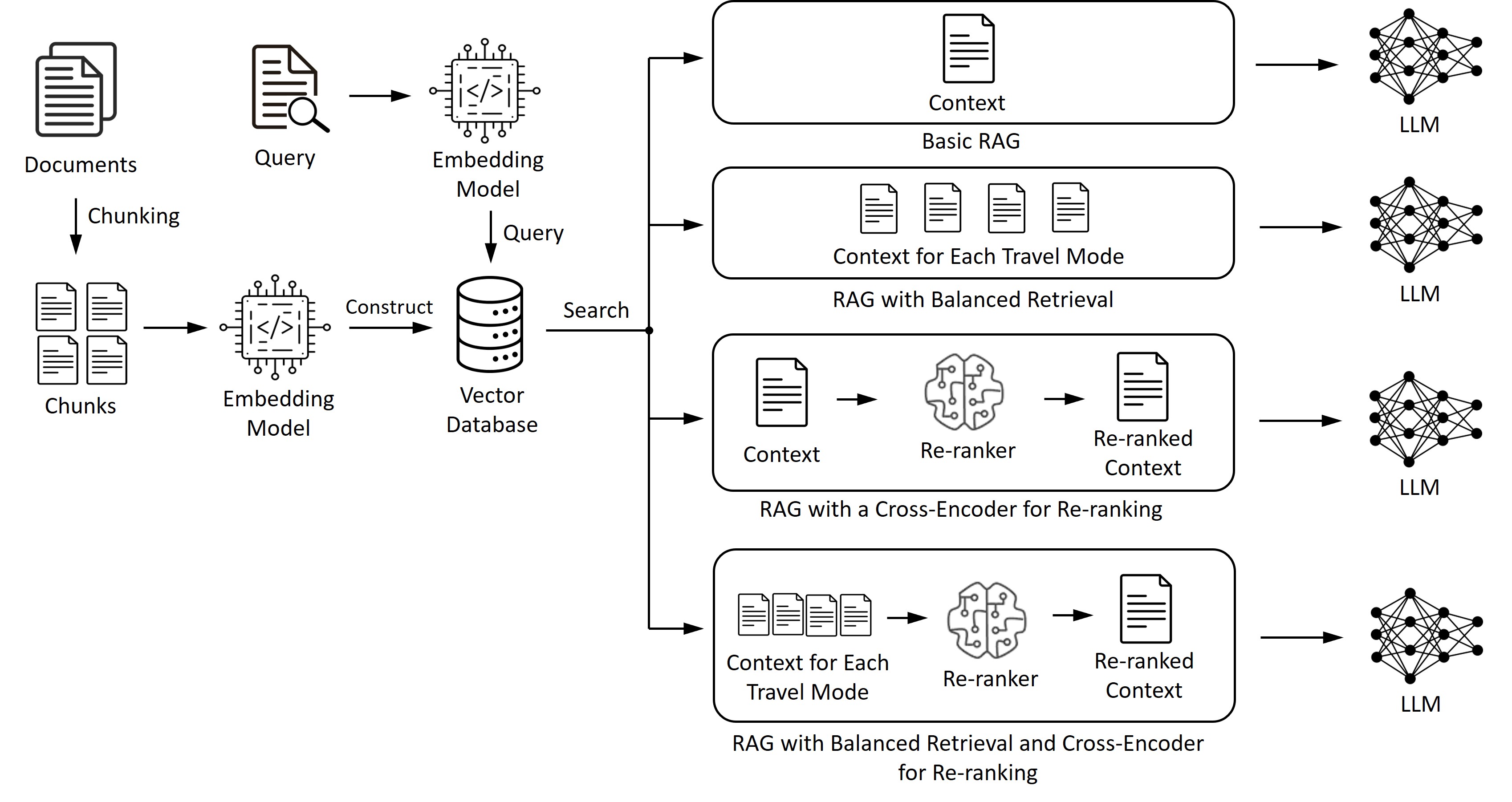} 
\caption{Overview of Basic RAG, RAG with Balanced Retrieval, RAG with a Cross-Encoder for Re-ranking, and RAG with Balanced Retrieval and Cross-Encoder for Re-ranking}
\label{fig:rag}
\end{figure}

The effectiveness of a RAG system is fundamentally dependent on the quality, relevance, and contextual alignment of the information retrieved and supplied to the LLM as external knowledge. The retrieval strategy determines which examples from the knowledge base are selected to guide the model’s reasoning and prediction process. To identify the most effective configuration for travel mode choice prediction, this study implements and systematically compares four representative RAG strategies that capture the major design dimensions of current retrieval-augmented frameworks:
(1) Basic RAG, which directly retrieves the most similar samples based on embedding similarity;
(2) RAG with Balanced Retrieval, which ensures a more even representation of different travel modes in the retrieved context to mitigate class imbalance;
(3) RAG with a Cross-Encoder for Re-ranking, which refines initial retrieval results by jointly encoding queries and candidates for higher semantic precision; and
(4) RAG with Balanced Retrieval and Cross-Encoder for Re-ranking, which integrates both balanced sampling and semantic refinement for comprehensive optimization.
These strategies are representative of the prevailing approaches in retrieval-augmented language modeling, spanning the spectrum from simple embedding-based retrieval to re-ranking-enhanced and distribution-balanced retrieval. Figure~\ref{fig:rag} provides an overview of the workflows for these strategies.

\subsubsection{Basic RAG}

The Basic RAG strategy represents a standard implementation of the retrieval-augmented framework. In this approach, the retriever leverages a vector store (e.g., FAISS) to perform semantic similarity search over an indexed corpus of training trip descriptions.  

Let each serialized trip description $d_i$ in the training dataset $\mathcal{D} = \{d_1, d_2, \dots, d_N\}$ be mapped into a high-dimensional embedding space $\mathbb{R}^m$ using a pre-trained embedding model $f_\theta(\cdot)$:
\begin{equation}
    \mathbf{z}_i = f_\theta(d_i) \in \mathbb{R}^m
\end{equation}

\noindent where $f_\theta$ is typically a transformer-based encoder (e.g., Sentence-BERT) parameterized by $\theta$. For a query trip description $q$ from the test set, the embedding is computed as:
\begin{equation}
    \mathbf{z}_q = f_\theta(q)
\end{equation}

To retrieve relevant context, the retriever computes the cosine similarity between the query embedding $\mathbf{z}_q$ and all stored embeddings $\mathbf{z}_i$ in the database:
\begin{equation}
    \text{sim}(\mathbf{z}_q, \mathbf{z}_i) = \frac{\mathbf{z}_q \cdot \mathbf{z}_i}{\|\mathbf{z}_q\|\|\mathbf{z}_i\|}
\end{equation}

The retriever then selects the top-$k$ documents:
\begin{equation}
    \mathcal{R}(q) = \underset{d_i \in \mathcal{D}}{\text{arg top-}k} \ \text{sim}(\mathbf{z}_q, \mathbf{z}_i)
\end{equation}

\noindent yielding a set of $k$ nearest neighbors $\{d_{(1)}, d_{(2)}, \dots, d_{(k)}\}$.  

These top-$k$ documents are concatenated to form an augmented context $\tilde{C}_q$, which is prepended to the query and passed to the LLM for conditional generation:

\begin{equation}
    y_q = \text{LLM}([q \oplus \tilde{C}_q])
\end{equation}

\noindent where $\oplus$ denotes concatenation and $y_q$ is the predicted outcome (i.e., travel mode).  

While this method is computationally efficient and straightforward to implement, it is limited by its reliance on pure semantic similarity. Specifically, high cosine similarity in the embedding space does not necessarily guarantee the retrieval of the most informative or task-relevant documents, potentially reducing prediction accuracy when the semantic neighbors are not aligned with the decision boundary relevant to the task.

\subsubsection{RAG with Balanced Retrieval}
Travel mode choice datasets are often characterized by a significant class imbalance, where common modes like driving are heavily overrepresented compared to minority modes like transit. A basic RAG approach on such a dataset is likely to retrieve a disproportionate number of examples from the majority class, leading to bias in the LLM's prediction. The RAG with Balanced Retrieval strategy is designed specifically to mitigate this issue. Instead of retrieving the top-k most similar documents overall, this method ensures a diverse and representative set of examples is selected. It works by performing a separate similarity search for each travel mode category and retrieving a pre-defined number of the top examples from each class. 

Let the set of possible travel mode classes be $\mathcal{C} = \{c_1, c_2, \dots, c_M\}$,
where $M$ denotes the number of travel mode categories. Each training document 
$d_i \in \mathcal{D}$ is associated with both its embedding $\mathbf{z}_i = f_\theta(d_i)$ 
and its mode label $\ell_i \in \mathcal{C}$. For a given query $q$, the embedding is computed as:  
\begin{equation}
    \mathbf{z}_q = f_\theta(q)
\end{equation}

Instead of a single global similarity ranking, Balanced Retrieval partitions the training 
corpus into subsets by class:  
\begin{equation}
    \mathcal{D}_{c_j} = \{ d_i \in \mathcal{D} \mid \ell_i = c_j \}, \quad j=1, \dots, M
\end{equation}

For each class $c_j$, the retriever selects the top-$k_j$ most similar documents based on cosine similarity:  
\begin{equation}
    \mathcal{R}_{c_j}(q) = \underset{d_i \in \mathcal{D}_{c_j}}{\text{arg top-}k_j} \ \text{sim}(\mathbf{z}_q, \mathbf{z}_i).
\end{equation}

The final retrieval set is the union of class-specific retrievals:  
\begin{equation}
    \mathcal{R}(q) = \bigcup_{j=1}^{M} \mathcal{R}_{c_j}(q),
\end{equation}

For balanced retrieval, $k_j$ is typically uniform across classes. {\YX For example, if $K=4$ and $M=4$, then the strategy retrieves one document per class ($k_1=k_2=k_3=k_4=1$), regardless of their prevalence in the dataset.}  
This balanced formulation ensures that the LLM has access to precedents from every possible travel mode, preventing majority classes from dominating the prompt. 

\subsubsection{RAG with a Cross-Encoder for Re-ranking}

The RAG with a Cross-Encoder for Re-Ranking strategy introduces a two-stage retrieval pipeline designed to maximize the relevance of the retrieved documents, thereby enhancing the quality of the context provided to the LLM. This approach balances computational efficiency with precision by combining a lightweight bi-encoder retriever for initial recall with a more expressive cross-encoder model for refined ranking.

In the first stage, a fast bi-encoder retriever is employed to identify a large set of candidate documents. Each document $d_i \in \mathcal{D}$ is encoded independently into a dense vector embedding 
$\mathbf{z}_i = f_\theta(d_i)$, while the query $q$ is embedded as $\mathbf{z}_q = f_\theta(q)$. 
The semantic similarity between query and document is measured using cosine similarity, and the retriever selects the top-$K'$ candidate documents:  
\begin{equation}
    \mathcal{C}(q) = \underset{d_i \in \mathcal{D}}{\text{arg top-}K'} \ \text{sim}(\mathbf{z}_q, \mathbf{z}_i)
\end{equation}

\noindent where $K'$ is larger than the final retrieval size ({\YX e.g., $K' = 20$}), ensuring broad coverage of potentially relevant candidates.  

{\YX
In the second stage, the candidate set $\mathcal{C}(q)$ is refined using a cross-encoder model $g_\phi(\cdot, \cdot)$, parameterized by $\phi$. Unlike the bi-encoder, which computes independent embeddings, the cross-encoder jointly processes the query and each candidate document as a pair $(q, d_i)$. The cross-encoder jointly encodes the query and candidate document as a paired input:
\begin{equation}
    x_i = [\texttt{CLS}] \ q \ [\texttt{SEP}] \ d_i \ [\texttt{SEP}]
\end{equation}
where \texttt{[CLS]} is the classification token and \texttt{[SEP]} is the separate token. The paired input is then passed through the MiniLM encoder \citep{wang2020minilm}. The hidden representation of the \texttt{[CLS]} token is used by the sequence-classification head to produce a scalar relevance score:
\begin{equation}
    \mathbf{h}_{\texttt{[CLS]},i}
    =
    \mathrm{MiniLM}_{\phi}(x_i)_{\texttt{[CLS]}}
\end{equation}
\begin{equation}
    s_i = g_{\phi}(q,d_i)
        = \mathbf{w}^{\top}\mathbf{h}_{\texttt{[CLS]},i} + b
\end{equation}
where $\mathbf{w}$ and $b$ are the parameters of the sequence-classification head, and $s_i$ is a scalar relevance score used to rank candidate examples.

The cross-encoder’s joint representation enables deeper semantic alignment, as it attends simultaneously to both query and document tokens. The candidates are then re-ranked according to their scores:  
\begin{equation}
    \mathcal{R}(q) = \underset{d_i \in \mathcal{C}(q)}{\text{arg top-}K} \ s_i
\end{equation}

\noindent where $K < K'$ specifies the final number of documents selected.  

Since the query and candidate example are encoded jointly, the cross-encoder can capture token-level interactions between the target trip and the retrieved example. This makes it more computationally expensive than bi-encoder retrieval, but it can provide more accurate relevance ranking for a relatively small candidate pool. By combining a fast approximate retriever for broad candidate selection with a computationally intensive cross-encoder for fine-grained ranking, this two-stage strategy ensures that the context provided to the LLM is both comprehensive and of the highest possible quality.  

In this study, the re-ranking module uses the pre-trained Sentence-Transformers cross-encoder model \texttt{ms-marco-MiniLM-L6-v2}. This model is a MiniLM-based BERT sequence-classification model trained on the MS MARCO Passage Ranking task \citep{reimers2019sentence}. We used the model off-the-shelf as a relevance scorer and did not conduct additional fine-tuning or task-specific training on the travel survey data. We selected this cross-encoder model because it provides a strong balance between re-ranking effectiveness and computational efficiency. Compared with larger cross-encoder variants, the MiniLM-L6 model is lightweight while maintaining competitive retrieval performance \citep{wang2020minilm}, making it suitable for repeated re-ranking over many test samples in the travel mode prediction experiment.

During inference, the cross-encoder was used in evaluation mode with fixed model parameters. Input pairs were tokenized with padding and truncation, with a maximum input length of 512 tokens. The raw scalar output of the sequence-classification head was used as the relevance score for ranking. If two candidates received the same score, ties were resolved deterministically by preserving the original ordering from the first-stage bi-encoder retrieval.

}

\subsubsection{RAG with Balanced Retrieval and Cross-Encoder for Re-Ranking}

The RAG with Balanced Retrieval and Cross-Encoder for Re-Ranking strategy integrates the complementary advantages of class balancing and fine-grained re-ranking into a unified two-stage pipeline. While Balanced Retrieval mitigates class imbalance by ensuring that minority modes are equally represented in the retrieved set, the Cross-Encoder introduces a subsequent refinement step to optimize contextual relevance. This combined approach ensures that the LLM receives in-context examples that are both diverse across classes and maximally relevant to the query.  

As in the Balanced Retrieval framework, the corpus is first partitioned into subsets by mode category:  
\begin{equation}
    \mathcal{D}_{c_j} = \{ d_i \in \mathcal{D} \mid \ell_i = c_j \}, \quad j=1, \dots, M
\end{equation}

\noindent where $M$ is the number of mode categories. 

From each class subset, the retriever selects the top-$k'_j$ candidates:  
\begin{equation}
    \mathcal{C}_{c_j}(q) = \underset{d_i \in \mathcal{D}_{c_j}}{\text{arg top-}k'_j} \ \text{sim}_{\text{bi}}(\mathbf{z}_q, \mathbf{z}_i)
\end{equation}

The union of these class-specific candidate sets forms the overall candidate pool:  
\begin{equation}
    \mathcal{C}(q) = \bigcup_{j=1}^{M} \mathcal{C}_{c_j}(q)
\end{equation}

The candidate pool $\mathcal{C}(q)$ is passed to a cross-encoder model $g_\phi(\cdot, \cdot)$, which jointly encodes the query--document pairs and produces contextualized relevance scores:  
\begin{equation}
    s_i = g_\phi(q, d_i), \quad d_i \in \mathcal{C}(q)
\end{equation}

The candidates are then re-ranked based on these scores, and the top-$K$ documents are selected:  
\begin{equation}
    \mathcal{R}(q) = \underset{d_i \in \mathcal{C}(q)}{\text{arg top-}K} \ s_i
\end{equation}

By combining balanced retrieval with cross-encoder re-ranking, this hybrid strategy ensures that retrieved examples are simultaneously representative across all classes and fine-tuned for semantic relevance, thereby achieving robustness against class imbalance and maximizing prediction quality.

\section{Experiments}
\label{s:4}
\subsection{Dataset}
The data for this study is sourced from the 2023 Puget Sound Regional Council (PSRC) Household Travel Survey. The PSRC survey is a comprehensive, ongoing data collection effort that captures day-to-day travel information from households across the central Puget Sound region in Washington State. 
The survey's primary goal is to develop a complete picture of regional travel patterns, including how, where, and why people travel, to support transportation and land-use planning. 
This rich dataset provides detailed records on individual trips, along with the socio-demographic characteristics of the travelers and their households. 

In this study, we extracted a specific subset of the survey data focusing on trips that occurred within the city of Seattle, WA. The temporal scope was limited to a single month, from May 1, 2023, to May 31, 2023, to ensure a consistent and contemporary sample of trips. The raw data underwent a cleaning process, which involved removing records with missing or incomplete information for the key variables required by the model. The final clean dataset consists of 2,847 individual trip records suitable for analysis. The dataset was then partitioned, with 80\% of the data allocated to the training set and the remaining 20\% reserved for the test set to evaluate model performance.

The processed dataset contains a set of independent variables selected based on their established influence on travel mode choice in the literature, as well as a single outcome variable representing the chosen mode. The outcome variable for the prediction task is Trip Mode, which is a categorical variable with four possible outcomes: Drive, Walk, Transit, and Bike/Micromobility. The independent variables used to model the travel mode choice are categorized as traveler characteristics and trip characteristics. Traveler characteristics describe the individual making the trip and their household context, including age, gender, education level, household income, and the number of vehicles in the household. Trip characteristics describe the trip itself, including the trip distance, the trip purpose (e.g., work, shopping, social), and the start time of the trip.
It is worth noting that the original dataset also contained a trip duration variable, which is a direct outcome of mode choice (e.g., walking a mile takes longer than driving). It would artificially inflate the model's predictive accuracy, compromising the validity of the evaluation. Therefore, the trip duration variable is excluded from the data to prevent information leakage. The descriptive statistics of the variables used in this study are presented in Table~\ref{tab:var_stat}.

\begin{table}[!ht]
\caption{Descriptive statistics of variables}
\label{tab:var_stat}
\renewcommand{\arraystretch}{1.1}
\setlength{\tabcolsep}{12pt} 
\begin{adjustbox}{width=0.8\textwidth,center}
\begin{tabular}{llll}
\Xhline{1.2pt}
\textbf{Variable}     & \textbf{Value}                & \textbf{Frequency/Mean} & \textbf{Percentage/SD} \\ \hline
\multicolumn{4}{l}{\textbf{Outcome Variable}}                                                            \\
Trip Mode             & Drive                         & 1271                    & 44.64\%                \\
                      & Walk                          & 1031                    & 36.21\%                \\
                      & Transit                       & 379                     & 13.31\%                \\
                      & Bike/Mircomobility            & 166                     & 5.83\%                 \\ \hline
\multicolumn{4}{l}{\textbf{Traveler   Characteristics}}                                                  \\
Age                   & 18-24                         & 288                     & 10.12\%                \\
                      & 25-34                         & 877                     & 30.80\%                \\
                      & 35-44                         & 757                     & 26.59\%                \\
                      & 45-54                         & 358                     & 12.57\%                \\
                      & 55-64                         & 276                     & 9.69\%                 \\
                      & 65-74                         & 224                     & 7.87\%                 \\
                      & 75 or older                   & 67                      & 2.35\%                 \\
Gender                & Male                          & 1261                    & 44.29\%                \\
                      & Female                        & 1530                    & 53.74\%                \\
                      & Non-binary                    & 56                      & 1.97\%                 \\
Education Level       & Less   than high school       & 25                      & 0.88\%                 \\
                      & High   school                 & 151                     & 5.30\%                 \\
                      & Some college                  & 239                     & 8.39\%                 \\
                      & Vocational/technical training & 14                      & 0.49\%                 \\
                      & Associates degree             & 128                     & 4.50\%                 \\
                      & Bachelor degree               & 1259                    & 44.22\%                \\
                      & Graduate degree               & 1031                    & 36.21\%                \\
Household Income      & Under \$25,000                & 182                     & 6.39\%                 \\
                      & \$25,000-\$49,999               & 447                     & 15.70\%                \\
                      & \$50,000-\$74,999               & 335                     & 11.77\%                \\
                      & \$75,000-\$99,999               & 465                     & 16.33\%                \\
                      & \$100,000-\$199,999             & 848                     & 29.79\%                \\
                      & \$200,000 or more             & 570                     & 20.02\%                \\
Vehicles in Household &                               & 1.03                    & 0.78                   \\ \hline
\multicolumn{4}{l}{\textbf{Trip Characteristics}}                                                        \\
Trip Distance (mile)  &                               & 2.28                    & 2.33                   \\
Start Time            &                               & 14.08                   & 4.32                   \\
Trip Purpose          & Home                          & 1074                    & 37.72\%                \\
                      & Work                          & 390                     & 13.70\%                \\
                      & Social/Recreation             & 463                     & 16.26\%                \\
                      & Shopping                      & 341                     & 11.98\%                \\
                      & Meal                          & 219                     & 7.69\%                 \\
                      & Business/Errand               & 140                     & 4.92\%                 \\
                      & Escort                        & 131                     & 4.60\%                 \\
                      & Overnight                     & 68                      & 2.39\%                 \\
                      & School                        & 18                      & 0.63\%                 \\
                      & Change mode                   & 3                       & 0.11\%                \\ \Xhline{1.2pt}
\end{tabular}
\end{adjustbox}
\end{table}

To assess the {\YX zero-shot transfer ability} of the models beyond the primary dataset, we utilized two additional datasets for evaluation. The first is a subset of the 2023 PSRC Household Travel Survey focusing specifically on trips that occurred within the city of Tacoma, WA. The second is a subset of the 2022 National Household Travel Survey (NHTS) restricted to trips that occurred within urban areas. From each of these two datasets, we randomly sampled 569 trips, which matches the sample size of our test set, to ensure a consistent evaluation setting. These additional datasets enable us to examine the transferability of the models to different contexts while controlling for sample size effects. The descriptive statistics of these test sets are presented in Table A.1 in the Appendix.

\subsection{Model Settings}
This section details the specific LLMs evaluated and the parameters used for both the LLM generation and the RAG pipeline. 

A key objective of this study is to understand the interplay between the base model's capabilities and the retrieval strategy. To this end, we selected three distinct LLM architectures from OpenAI, representing different points on the spectrum of capability, speed, and cost:
\begin{itemize}
    \item \textbf{GPT-4o}: OpenAI's multimodal model, which can process and generate combinations of text, audio, and images. Moderate speed, moderate reasoning ability, high cost. In this study, the temperature parameter was set to 0.
    \item \textbf{o3}: OpenAI's powerful reasoning model, which excels at complex tasks that require step-by-step logical deduction. Slow speed, high reasoning ability, high cost.
    \item \textbf{o4-mini}: A smaller, faster, and more affordable reasoning model, optimized for high efficiency and speed, particularly in reasoning tasks. Fastest and cheapest.
\end{itemize}

{\YX
The configuration of RAG was held constant across all three LLMs. We used OpenAI's \texttt{text-embedding-3-large} model as the embedding model. The vector embeddings of the training data were indexed and stored using FAISS. The FAISS index was implemented using \texttt{IndexFlatL2}, and semantic similarity was measured by cosine similarity. All models and retrieval strategies shared the same training-set vector store to ensure comparability. The cross-encoder re-ranking model was \texttt{ms-marco-MiniLM-L6-v2}.

In all experiments, the final number of retrieved documents (i.e., trip examples) included in the LLM prompt was fixed at $K=4$. For Basic RAG, the retriever selected the top four examples based on embedding similarity. For RAG with Balanced Retrieval, we set $k_j=1$ for each class $j$, so that one example was retrieved from each travel mode category and the final retrieval set contained $K=\sum_{j=1}^{4}k_j=4$ examples.
For RAG with Cross-Encoder Re-ranking, we used a two-stage retrieval process. The initial bi-encoder retriever selected $K'=20$ candidate examples based on embedding similarity. The cross-encoder then re-ranked these candidates, and the top $K=4$ examples were included in the final prompt. For RAG with Balanced Retrieval and Cross-Encoder Re-ranking, the initial candidate pool was also set to $K'=20$, but candidates were retrieved in a class-balanced manner, with $k'_j=5$ examples retrieved from each of the four travel mode classes. These 20 candidates were then re-ranked by the cross-encoder, and the final top $K=4$ examples were used as in-context examples.

The retrieval number $K=4$ was selected as a compact and task-appropriate setting to balance retrieval coverage and prompt efficiency. In our study, the prediction task involves four travel mode categories. Setting $K=4$ allows the retrieval module to provide a small but informative set of examples while keeping the prompt concise. For the balanced retrieval strategy, setting $K=4$ also corresponds to one retrieved example per class. 
In existing studies, the number of retrieved documents is usually set to a small value. For example, \cite{lewis2020retrieval} consider $k \in \{5, 10\}$ for training and set test-time $k$ using development data. Other studies have shown that retrieving too many passages can introduce noise, redundancy, and additional computational cost, and can even degrade generation quality \citep{gao2024retrieval, zhu2024information}. In addition, long or crowded contexts are not always used robustly by LLMs \citep{liu2024lost}. Therefore, we adopted a relatively small retrieval size to maintain a favorable balance between contextual support and prompt compactness. 
Similarly, we selected $K'=20$ as a practical trade-off between retrieval coverage and computational efficiency. A larger candidate pool may increase the chance of including highly relevant examples, but it also increases the computational cost of cross-encoder inference. Since the final prompt only includes $K=4$ documents, retrieving $K'=20$ candidates provides a fivefold candidate pool for re-ranking while keeping it computationally manageable. This setting also aligns with the balanced retrieval design, where four travel mode classes are represented by five candidates each before re-ranking. 

Ties in similarity or re-ranking scores were handled deterministically using the stable ordering returned by the retrieval index. If a class contained fewer than the required number of examples, all available examples from that class were retained, and the remaining slots were filled using the highest-ranked unused examples from the global similarity ranking. In the main experiments, since all four travel mode classes were present in the training set, this fallback rule was rarely invoked. In some cases, a class contained more than the required number of training examples, but the retriever returned fewer than the target number for that class. We treated this as a retrieval shortfall rather than a class-size shortfall. When this occurred, we first supplemented the missing examples using the highest-ranked unused examples from the same class whenever available. If the same-class supplementation still did not provide enough examples, the remaining slots were filled using the highest-ranked unused examples from the global retrieval list.
}

\subsection{Baseline Models}

To properly evaluate the performance of the proposed frameworks, we selected four baseline models, {\YX including Multinomial Logit (MNL), Random Forest (RF), Multi-layer Perceptron (MLP), and Zero-Shot LLM. The conventional baseline models (i.e., MNL, RF, and NLP) were trained using the same training set used to construct the RAG knowledge base. Categorical variables were encoded consistently across models. Hyperparameters were selected using grid search and cross-validation on the training set. The final conventional baseline models were then refitted on the full training set using the selected hyperparameters. The models were evaluated on the same test set used for the LLM-based models. For the two external datasets, the trained baseline models were directly evaluated without additional hyperparameter tuning, consistent with the evaluation protocol used for the LLM-based approaches. The details of the baseline models are as follows.}
\begin{itemize}
    \item \textbf{Multinomial Logit (MNL)}: MNL is the most widely-used discrete choice model in transportation research. It assumes that each alternative has an associated utility composed of a deterministic part and a random error. The choice probabilities are calculated based on utility. {\YX After hyperparameter tuning, the inverse of regularization strength C is set to 1; the penalty is set to L2; and the solver is L-BFGS.}
    \item \textbf{Random Forest (RF)}: RF is an ensemble learning method built on decision trees, widely recognized for its strong performance in travel mode choice modeling. Its data-driven, non-parametric structure enables it to capture complex, non-linear relationships within the data, leading to robust and accurate predictions. {\YX After hyperparameter tuning, the number of trees is set to 80; the maximum depth of the tree is set to 10; the minimum number of samples required to split a node is set to 2; the number of features to consider when looking for the best split is set to “sqrt”. }
    \item \textbf{Multi-layer Perceptron (MLP)}: MLP is well-regarded for its ability to extract and model complex, high-dimensional patterns directly from data. MLPs have been widely applied in transportation research for various classification problems. {\YX After hyperparameter tuning, the hidden layer size is set to (32,32); the activation is ReLU; the solver is Adam; the alpha is set to 0.0001; and the maximum number of iterations is set to 1000. }
    \item \textbf{Zero-Shot LLM}: In this configuration, the LLM is prompted with only a description of the task and the serialized text of the trip to be predicted, without being provided any in-context examples from the training data. This method evaluates the LLM's inherent ability to perform the travel mode choice prediction task based solely on its pre-trained knowledge and reasoning capabilities. 
\end{itemize}

\subsection{Evaluation}
We employ four metrics to evaluate the predictive performance of the models, including accuracy, precision, recall, and the F1-score. The metrics are calculated by:	
\begin{align}
    Accuracy&=\frac{TP+TN}{TP+TN+FP+FN} \\
    Precision&=\frac{TP}{TP+FP} \\
    Recall&=\frac{TP}{TP+FN}\\
    F1&=\frac{2\times Precision\times Recall}{Precision+Recall}
\end{align}

\noindent where TP is true positives, TN is true negatives, FP is false positives, and FN is false negatives. For this multi-class problem, we use weighted precision, recall, and F1-scores, where each class’s metric is weighted by its sample size.

\section{Results}
\label{s:5}

\subsection{Model Performance}
\begin{figure}[!ht]
\centering
\includegraphics[width=\textwidth]{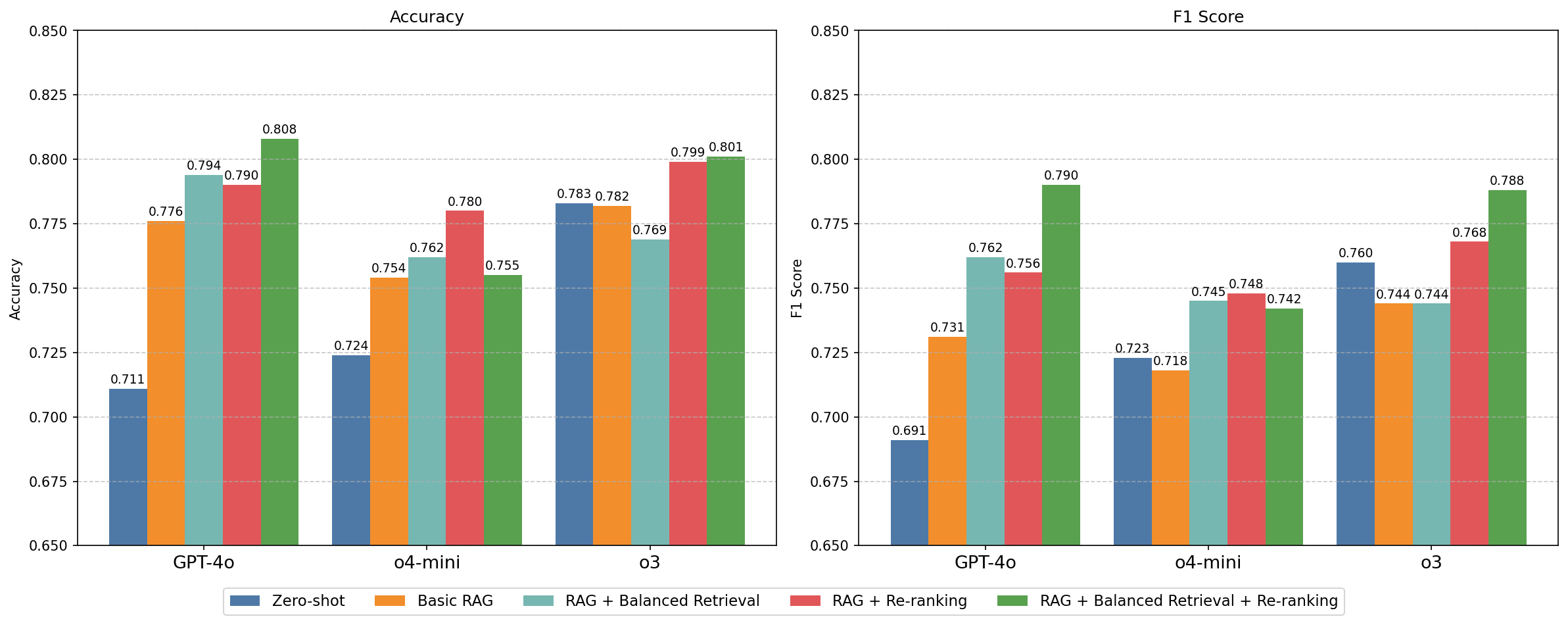} 
\caption{Model performance with different LLM architectures and RAG strategies}
\label{fig:performance}
\end{figure}

The comprehensive empirical outcomes of our comparative evaluation are presented in Table~\ref{tab:performance}. The highest overall classification accuracy achieved is 0.808, obtained by the GPT-4o model when augmented with the RAG with Balanced Retrieval and Cross-Encoder for Re-ranking approach. This combination also yielded the highest F1 score (0.790) and recall (0.808). Concurrently, the highest measure of precision is 0.804, realized by the GPT-4o model, through the application of RAG with a Cross-Encoder for Re-ranking. Notably, all RAG-augmented LLMs outperformed the statistical and machine learning baselines, underscoring the effectiveness of RAG in enhancing LLM performance in our experiments. The performance of models with different LLM architectures and RAG strategies is presented in Figure~\ref{fig:performance}.

\begin{table}[!ht]
\caption{Performance of baseline models and LLMs under different RAG strategies. The best-performing value is highlighted in \textbf{bold}, and the second-best value is \underline{underlined}.}
\label{tab:performance}
\renewcommand{\arraystretch}{1.3} 
\setlength{\tabcolsep}{10pt} 
\centering
\begin{adjustbox}{width=0.9\textwidth,center}
\begin{tabular}{p{2cm}p{5cm}ccccc}
\Xhline{1.2pt} 
\textbf{Model}           & \textbf{Method}          & \textbf{Accuracy} & \textbf{F1 Score} & \textbf{Recall} & \textbf{Precision} \\ \hline
MNL                      & -                                       & 0.738    & 0.698    & 0.738  & 0.678     \\
RF                       & -                                       & 0.747    & 0.714    & 0.747  & 0.691     \\
MLP                      & -                                       & 0.736    & 0.716    & 0.736  & 0.709     \\ \hline
\multirow{5}{*}{GPT-4o}  & Zero-shot                               & 0.711    & 0.691    & 0.711  & 0.737     \\
                         & Basic RAG                               & 0.776    & 0.731    & 0.776  & 0.727     \\
                         & RAG + Balanced   Retrieval              & 0.794    & 0.762    & 0.794  & \underline{0.801}     \\
                         & RAG +   Re-ranking                      & 0.790    & 0.756    & 0.790  & \textbf{0.804}     \\
                         & RAG + Balanced   Retrieval + Re-ranking & \textbf{0.808}    & \textbf{0.790}    & \textbf{0.808}  & 0.801     \\ \hline
\multirow{5}{*}{o4-mini} & Zero-shot                               & 0.724    & 0.723    & 0.724  & 0.737     \\
                         & Basic RAG                               & 0.754    & 0.718    & 0.754  & 0.732     \\
                         & RAG + Balanced   Retrieval              & 0.762    & 0.745    & 0.762  & 0.739     \\
                         & RAG +   Re-ranking                      & 0.780    & 0.748    & 0.780  & 0.747     \\
                         & RAG + Balanced   Retrieval + Re-ranking & 0.755    & 0.742    & 0.755  & 0.743     \\ \hline
\multirow{5}{*}{o3}      & Zero-shot                               & 0.783    & 0.760    & 0.783  & 0.754     \\
                         & Basic RAG                               & 0.782    & 0.744    & 0.782  & 0.757     \\
                         & RAG + Balanced   Retrieval              & 0.769    & 0.744    & 0.769  & 0.750     \\
                         & RAG +   Re-ranking                      & 0.799    & 0.768    & 0.799  & 0.792     \\
                         & RAG + Balanced   Retrieval + Re-ranking & \underline{0.801}    & \underline{0.788}    & \underline{0.801}  & 0.786     \\ \Xhline{1.2pt} 
\end{tabular}
\end{adjustbox}
\end{table}

A significant finding is the exceptionally strong zero-shot performance of the o3 model, which achieved an accuracy of 0.783 and an F1 score of 0.760. This capability surpasses all traditional baselines and even the RAG-augmented performance of the o4-mini model. This high baseline fundamentally alters the impact of RAG for this specific model. While the introduction of a basic RAG pipeline provided a solid improvement for GPT-4o (accuracy increased from 0.711 to 0.776) and o4-mini (accuracy increased from 0.724 to 0.754), it had an adverse effect on the o3 model. The accuracy of o3 model slightly decreased to 0.782 with Basic RAG, suggesting that for a model with an advanced reasoning capability, a simple retrieval mechanism can introduce noise that hinders rather than helps performance.

This trend continues with the RAG with Balanced Retrieval. While this technique improved performance for both GPT-4o (accuracy increased from 0.711 to 0.794) and o4-mini (accuracy increased from 0.724 to 0.762), it caused a further decline in performance for the o3 model, with accuracy dropping from 0.783 to 0.769. This reinforces the observation that for the o3 model, standard and moderately improved RAG techniques were not only failing to unlock further potential but were actively degrading its high-quality zero-shot predictions. For GPT-4o, however, this method was highly effective, boosting accuracy from 0.711 to 0.794 and F1 score from 0.691 to 0.762.

The RAG with a Cross-Encoder for Re-ranking strategy proved to be effective in boosting performance, though its effect varied by model. For the o3 model, this RAG method provided a benefit over its zero-shot baseline, pushing its accuracy to 0.799. This indicates that only a highly optimized retrieval pipeline that surfaces the most relevant evidence is capable of augmenting o3's powerful intrinsic reasoning. For GPT-4o, the re-ranker acted as a specialist tool, slightly lowering accuracy from its peak but increasing precision to 0.804, the highest recorded value. This highlights a critical trade-off: the o3 configuration is superior for overall accuracy, while the GPT-4o setup is preferable for applications where avoiding false positives is paramount.

The combination of Balanced Retrieval with Cross-Encoder Re-Ranking yielded the strongest overall performance for the GPT-4o model and the o3 model, underscoring the complementary strengths of diversity and precision in retrieval. For GPT-4o, this hybrid strategy proved particularly effective, achieving an accuracy of 0.808, an F1 score of 0.790, and a recall of 0.808, which are the highest values observed in the entire evaluation. In addition, this improvement was obtained without sacrificing precision, which remained competitive at 0.801. For the o3 model, the combined strategy also reached its peak performance with an accuracy of 0.801 and F1 score of 0.788, thereby surpassing its already strong zero-shot and re-ranking-only baselines. By contrast, o4-mini experienced only marginal gains from the combination. These results indicate that while advanced models like GPT-4o and o3 can benefit from re-ranking alone, the fusion of class-balancing and re-ranking is the most robust retrieval strategy.

In summary, the results highlight the complex interplay between LLM architecture and RAG strategy. While all RAG-augmented configurations outperform traditional baselines, the path to optimal performance differs across models. The o3 model demonstrates strong zero-shot performance, with basic and balanced retrieval methods actually degrading its accuracy, emphasizing that retrieval augmentation is not inherently beneficial for models with advanced reasoning capabilities. Nevertheless, when paired with a highly optimized pipeline that combines balanced retrieval with cross-encoder re-ranking, o3 achieves its peak performance, surpassing its zero-shot baseline in accuracy, F1 score, and recall. For GPT-4o, the hybrid strategy delivers the best overall results, producing the highest observed accuracy (0.808), F1 score (0.790), and recall (0.808), while re-ranking alone yields the strongest precision (0.804). These findings demonstrate that effective augmentation requires alignment between model architecture and retrieval design.

{\YX
To further examine the influence of class imbalance, we present the accuracy and F1-score for each travel mode category in Table~\ref{tab:mode_performance}. The results show that most models perform well for the majority classes (i.e., Drive and Walk), but exhibit weaker performance for the less frequent classes (i.e., Transit and Bike/Micromobility). This pattern is especially evident for the traditional baseline models, which achieve high F1-scores for Drive and Walk but perform poorly for Bike/Micromobility.

In contrast, the LLM-based models demonstrate improved performance for minority modes, particularly when retrieval strategies are designed to account for class imbalance. The combination of Balanced Retrieval and Cross-Encoder Re-ranking yields the strongest overall minority-class performance. For Transit, this strategy achieves F1-scores of 0.574 for GPT-4o, 0.488 for o4-mini, and 0.576 for o3. For Bike/Micromobility, this strategy improves the F1-score to 0.293 for GPT-4o, 0.291 for o4-mini, and 0.304 for o3, outperforming both the traditional baselines and simpler RAG configurations. These results suggest that balanced retrieval helps expose the LLM to examples from underrepresented modes, while re-ranking improves the contextual relevance of the retrieved examples. Therefore, the proposed RAG framework can partially mitigate the effects of dataset imbalance, although predicting the rare modes such as Bike/Micromobility remains challenging.

}

\begin{table}[!ht]
\caption{\YX Model performance for each travel mode category.}
\label{tab:mode_performance}
\renewcommand{\arraystretch}{1.3} 
\setlength{\tabcolsep}{6pt} 
\centering
\begin{adjustbox}{width=\textwidth,center}
{\YX
\begin{tabular}{p{2cm}p{5cm}cccccccc}
\Xhline{1.2pt} 

\multirow{2}{*}{\textbf{Model}} & \multicolumn{1}{c}{\multirow{2}{*}{\textbf{Method}}}                               & \multicolumn{4}{c}{\textbf{Accuracy}}                                                   & \multicolumn{4}{c}{\textbf{F1   Score}}                                                 \\
                                & \multicolumn{1}{c}{}                                                               & Drive & Walk  & Transit & \begin{tabular}[c]{@{}c@{}}Bike/\\ Micromobility\end{tabular} & Drive & Walk  & Transit & \begin{tabular}[c]{@{}c@{}}Bike/\\ Micromobility\end{tabular} \\ \hline
\multicolumn{1}{l}{MNL}         & -                                                                                  & 0.859 & 0.907 & 0.256   & 0.000                                                         & 0.813 & 0.818 & 0.342   & 0.000                                                         \\
\multicolumn{1}{l}{RF}          & -                                                                                  & 0.887 & 0.859 & 0.354   & 0.000                                                         & 0.824 & 0.815 & 0.423   & 0.000                                                         \\
\multicolumn{1}{l}{MLP}         & -                                                                                  & 0.827 & 0.863 & 0.427   & 0.061                                                         & 0.818 & 0.792 & 0.470   & 0.098                                                         \\ \hline
\multirow{5}{*}{GPT-4o}         & Zero-shot                                                                          & 0.704 & 0.966 & 0.276   & 0.182                                                         & 0.788 & 0.752 & 0.396   & 0.245                                                         \\
                                & Basic   RAG                                                                        & 0.929 & 0.908 & 0.250   & 0.030                                                         & 0.859 & 0.826 & 0.373   & 0.059                                                         \\
                                & RAG +   Balanced Retrieval                                                         & 0.897 & 0.932 & 0.408   & 0.030                                                         & 0.861 & 0.844 & 0.517   & 0.059                                                         \\
                                & RAG +   Re-ranking                                                                 & 0.905 & 0.927 & 0.368   & 0.030                                                         & 0.856 & 0.838 & 0.505   & 0.059                                                         \\
                                & \begin{tabular}[c]{@{}l@{}}RAG +   Balanced Retrieval \\ + Re-ranking\end{tabular} & 0.905 & 0.908 & 0.487   & 0.182                                                         & 0.876 & 0.844 & 0.574   & 0.293                                                         \\ \hline
\multirow{5}{*}{o4-mini}        & Zero-shot                                                                          & 0.751 & 0.913 & 0.342   & 0.212                                                         & 0.805 & 0.823 & 0.416   & 0.171                                                         \\
                                & Basic   RAG                                                                        & 0.941 & 0.820 & 0.250   & 0.061                                                         & 0.832 & 0.805 & 0.376   & 0.093                                                         \\
                                & RAG +   Balanced Retrieval                                                         & 0.913 & 0.796 & 0.447   & 0.121                                                         & 0.848 & 0.794 & 0.515   & 0.174                                                         \\
                                & RAG +   Re-ranking                                                                 & 0.913 & 0.893 & 0.355   & 0.030                                                         & 0.845 & 0.840 & 0.478   & 0.053                                                         \\
                                & \begin{tabular}[c]{@{}l@{}}RAG +   Balanced Retrieval \\ + Re-ranking\end{tabular} & 0.909 & 0.777 & 0.408   & 0.242                                                         & 0.826 & 0.806 & 0.488   & 0.291                                                         \\ \hline
\multirow{5}{*}{o3}             & Zero-shot                                                                          & 0.874 & 0.937 & 0.368   & 0.091                                                         & 0.865 & 0.846 & 0.452   & 0.133                                                         \\
                                & Basic   RAG                                                                        & 0.937 & 0.888 & 0.303   & 0.030                                                         & 0.849 & 0.839 & 0.434   & 0.056                                                         \\
                                & RAG +   Balanced Retrieval                                                         & 0.885 & 0.869 & 0.421   & 0.061                                                         & 0.844 & 0.806 & 0.516   & 0.108                                                         \\
                                & RAG +   Re-ranking                                                                 & 0.913 & 0.932 & 0.382   & 0.061                                                         & 0.864 & 0.850 & 0.513   & 0.111                                                         \\
                                & \begin{tabular}[c]{@{}l@{}}RAG +   Balanced Retrieval \\ + Re-ranking\end{tabular} & 0.909 & 0.864 & 0.526   & 0.212                                                         & 0.880 & 0.832 & 0.576   & 0.304                                                         \\ \hline
\end{tabular}
}
\end{adjustbox}
\end{table}

{\YX
\subsection{ Efficiency Analysis}

We further examined the computational efficiency of each model and RAG strategy combination. The results are presented in Table~\ref{tab:efficiency}. Specifically, we report the average API runtime, RAG runtime, total runtime, prompt token consumption, reasoning token consumption, total token consumption, and estimated API usage cost. 

\begin{table}[!ht]
\caption{\YX Computational cost of LLMs under different RAG strategies.}
\label{tab:efficiency}
\renewcommand{\arraystretch}{1.3} 
\setlength{\tabcolsep}{10pt} 
\centering
\begin{adjustbox}{width=\textwidth,center}
{\YX
\begin{tabular}{p{2cm}p{5cm}ccccccc}
\Xhline{1.2pt} 
\multirow{2}{*}{\textbf{Model}}                    & \multicolumn{1}{c}{\multirow{2}{*}{\textbf{Method}}}                  & \multicolumn{3}{c}{\textbf{Average Runtime (s)}}                                                 & \multicolumn{3}{c}{\textbf{Average Token Cost}}                                                 & \multirow{2}{*}{\begin{tabular}[c]{@{}c@{}}\textbf{API Usage} \\ \textbf{Cost (\$)}\end{tabular}}                 \\
                                          & \multicolumn{1}{c}{}                                         & API                         & RAG                         & Total                       & Prompt                     & Reasoning                     & Total                     &                                                                                                 \\ \hline
\multirow{5}{*}{GPT-4o}                   & Zero-shot                                                    & 0.481                       & 0.000                       & 0.481                       & 115.3                      & 0.0                           & 115.3                     & 0.0003                                                                                          \\
                                          & Basic   RAG                                                  & 0.463                       & 0.177                       & 0.639                       & 355.2                      & 0.0                           & 355.2                     & 0.0009                                                                                          \\
                                          & RAG +   Balanced Retrieval                                   & 0.424                       & 0.748                       & 1.171                       & 360.4                      & 0.0                           & 360.4                     & 0.0009                                                                                          \\
                                          & RAG +   Re-ranking                                           & 0.414                       & 0.229                       & 0.643                       & 355.4                      & 0.0                           & 355.4                     & 0.0009                                                                                          \\
                                          & RAG +   Balanced Retrieval + Re-ranking                      & 0.425                       & 0.832                       & 1.257                       & 360.8                      & 0.0                           & 360.8                     & 0.0009                                                                                          \\ \hline
\multirow{5}{*}{o4-mini}                  & Zero-shot                                                    & 2.198                       & 0.000                       & 2.198                       & 114.3                      & 111.7                         & 226.0                     & 0.0008                                                                                          \\
                                          & Basic   RAG                                                  & 2.549                       & 0.204                       & 2.753                       & 354.2                      & 152.7                         & 506.9                     & 0.0012                                                                                          \\
                                          & RAG +   Balanced Retrieval                                   & 2.742                       & 0.767                       & 3.508                       & 359.4                      & 209.8                         & 569.2                     & 0.0014                                                                                          \\
                                          & RAG +   Re-ranking                                           & 2.354                       & 0.221                       & 2.575                       & 354.4                      & 148.5                         & 502.9                     & 0.0012                                                                                          \\
                                          & RAG +   Balanced Retrieval + Re-ranking                      & 3.597                       & 0.901                       & 4.498                       & 359.8                      & 264.0                         & 623.8                     & 0.0017                                                                                          \\ \hline
\multirow{5}{*}{o3}                       & Zero-shot                                                    & 3.606                       & 0.000                       & 3.606                       & 114.3                      & 113.6                         & 227.9                     & 0.0015                                                                                          \\
                                          & Basic   RAG                                                  & 3.293                       & 0.201                       & 3.494                       & 354.2                      & 97.0                          & 451.2                     & 0.0018                                                                                          \\
                                          & RAG +   Balanced Retrieval                                   & 3.510                       & 0.786                       & 4.296                       & 359.6                      & 155.4                         & 515.0                     & 0.0020                                                                                          \\
                                          & RAG +   Re-ranking                                           & 3.013                       & 0.233                       & 3.246                       & 354.4                      & 99.9                          & 454.3                     & 0.0018                                                                                          \\
                                          & RAG +   Balanced Retrieval + Re-ranking                      & 5.393                       & 0.894                       & 6.287                       & 359.8                      & 221.1                         & 580.8                     & 0.0028                                                                                          \\ \Xhline{1.2pt}
\multicolumn{9}{l}{\begin{tabular}[c]{@{}l@{}}Note:   1. The observed execution times are cumulative of model processing and inherent API latency. While these \\values provide a baseline for comparison,   they are subject to network and server stochasticity. \\      2. API expenditures are calculated based on the OpenAI official pricing schedule as of April 2026.\end{tabular}}
\end{tabular}
}
\end{adjustbox}
\end{table}

The results show that GPT-4o is the most efficient model in terms of both latency and API cost. In the zero-shot setting, GPT-4o requires only 0.481 seconds and 115.3 tokens per sample, corresponding to an estimated cost of \$0.0003. Even under the most computationally intensive setting (i.e., RAG with Balanced Retrieval and Re-ranking), its total runtime remains 1.257 seconds, with 360.8 total tokens and an estimated cost of \$0.0009 per sample. By comparison, o4-mini requires 2.198 to 4.498 seconds and \$0.0008 to \$0.0017 per sample across settings, while o3 requires 3.013 to 6.287 seconds and \$0.0015 to \$0.0028 per sample. These results indicate that GPT-4o offers a more favorable efficiency profile than the o4-mini and o3 models.

RAG substantially increases token usage for all models and generally increases runtime as well. Across all three LLMs, zero-shot prompting uses about 114--115 prompt tokens per sample, whereas RAG-based methods increase prompt token usage to approximately 354--360 tokens because retrieved examples are appended to the prompt. Among the RAG strategies, balanced retrieval contributes the largest retrieval overhead, increasing the RAG runtime from about 0.177--0.233 seconds under Basic RAG or Re-ranking to about 0.748--0.901 seconds when balanced retrieval is used. In contrast, the total token cost remains relatively similar across different RAG strategies, since the retrieved context length is comparable.

For the reasoning models (i.e., o4-mini and o3), the additional reasoning tokens increase total token usage. For o4-mini, reasoning token consumption ranges from 111.7 tokens in zero-shot to 264.0 tokens in the combined balanced-retrieval-and-re-ranking setting. For o3, the corresponding range is 97.0--221.1 tokens under RAG-based settings, with 113.6 tokens in zero-shot.


In summary, while advanced retrieval pipelines can improve predictive performance, they also introduce additional computational overhead, especially when balanced retrieval is included. Nevertheless, the best-performing configuration in our main experiment, GPT-4o with Balanced Retrieval and Re-ranking, remains lightweight in absolute terms.
}

{\YX
\subsection{Prediction Error Analysis}

To better understand the sources of prediction error, we examined the confusion matrices for all model–method combinations based on the test set (Figure~\ref{fig:confusion}). The results show that the misclassifications are not evenly distributed across classes. Across nearly all settings, Drive and Walk are predicted much more accurately, while Transit and Bike/Micromobility are more likely to be misclassified.

\begin{figure}[!ht]
\centering
\includegraphics[width=\textwidth]{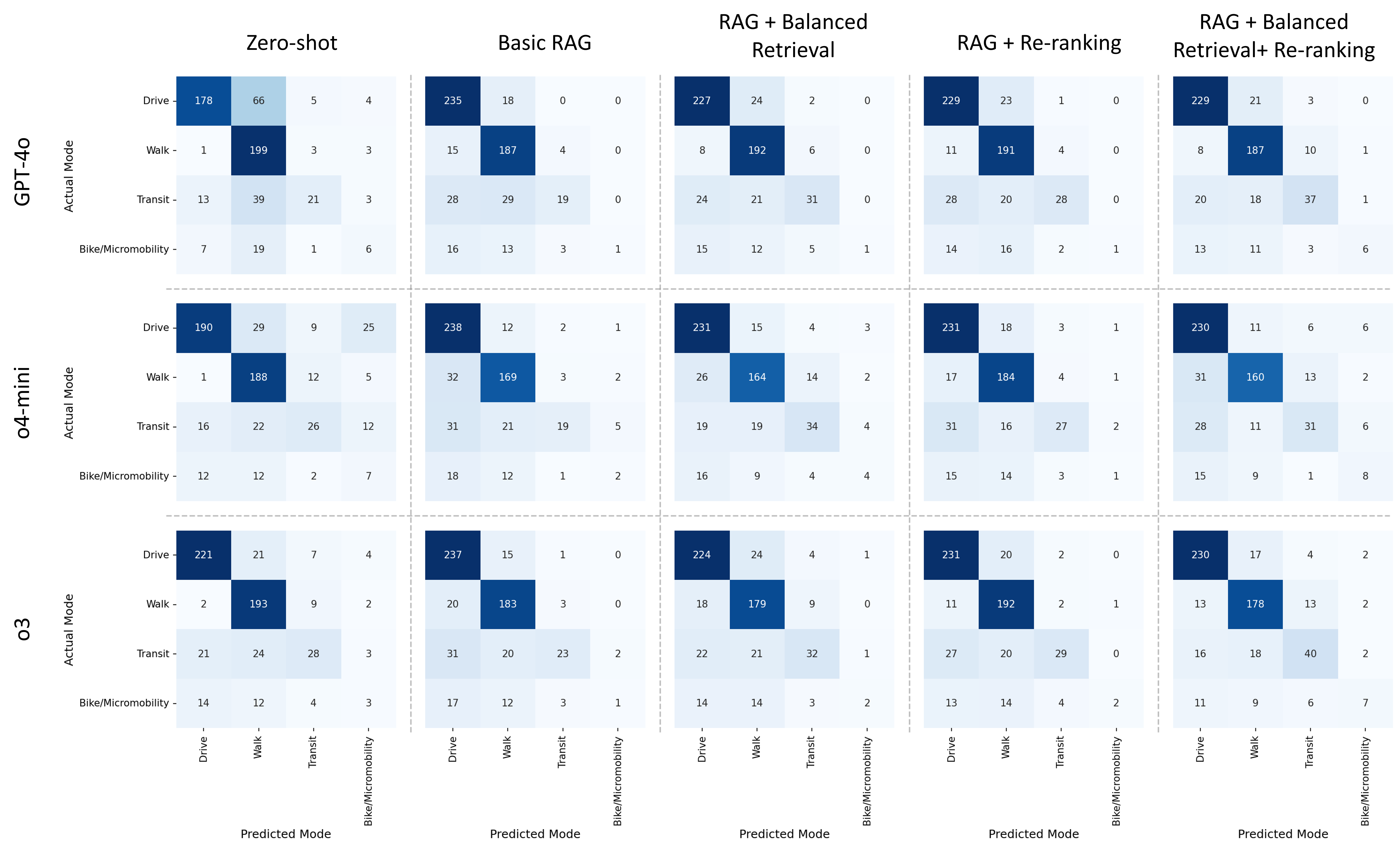}
\caption{\YX Confusion matrices of travel mode prediction for different LLMs and retrieval strategies based on the test set}
\label{fig:confusion}
\end{figure}

Specifically, Transit trips are most often confused with Drive or Walk. For example, in the GPT-4o zero-shot setting, only 21 of 76 transit samples are correctly classified, while 39 are predicted as Walk. Even in the best-performing overall configuration, GPT-4o with Balanced Retrieval and Re-ranking, while the number of correctly identified transit trips increases to 37, the remaining errors are still mainly in Drive and Walk. This result indicates that transit trips are behaviorally harder to separate from neighboring urban modes (i.e., drive and walk) when only traveler and trip descriptors are provided.

Bike/Micromobility is the most difficult class to predict overall. Across the confusion matrices, this class is frequently predicted as Drive or Walk, and its correct classification rate remains substantially lower than the other three classes. This pattern is consistent with the fact that Bike/Micromobility is the smallest class in the dataset and likely has the weakest representation in both model learning and retrieval.

There are three main reasons for these misclassification patterns. First, the dataset is imbalanced. Drive and Walk account for the majority of samples, while Transit and Bike/Micromobility are much less frequent. Second, the input variables mainly describe traveler and trip characteristics, but do not include alternative-specific level-of-service variables such as mode-specific travel time, cost, access to transit, or bicycle infrastructure. Without these attributes, Transit and Bike/Micromobility trips can resemble Drive or Walk trips. Third, the Bike/Micromobility class combines multiple relatively rare modes into one category, which increases within-class heterogeneity and makes this class harder to distinguish.

In addition, the confusion matrices also indicate that improved retrieval design can partly mitigate these errors. Balanced retrieval and re-ranking tend to improve recognition of minority classes, especially Transit, by providing more diverse and relevant in-context examples. 

}

\subsection{\YX Zero-Shot Transferability to External Datasets}

We used two external datasets to evaluate the {\YX model transferability beyond the primary dataset under a no-adaptation protocol.} These datasets include a subset of 2023 PSRC Household Travel Survey trips within Tacoma, WA, which is more similar in sampling and context to the training and test data, and a subset of 2022 NHTS trips within urban areas, which represents a more distinct distributional shift. The performance of the models on these datasets is presented in Table~\ref{tab:generalization}.

{\YX 
Importantly, this experiment is designed as a zero-shot transfer evaluation rather than a target-domain re-estimation one. All models were directly tested on the external datasets without any retraining, hyperparameter tuning, or recalibration using target-domain labels. The conventional baseline models were trained only on the source training set, while the LLM-based models were applied through zero-shot prompting or retrieval from the source-domain knowledge base. Therefore, the results should be interpreted as evidence of model robustness under a no-retraining deployment scenario, rather than as a comparison against conventional models that have been re-estimated for the target region.

This evaluation setting is practically meaningful because labeled travel survey data in a new region may be unavailable, limited, or costly to obtain \citep{mohammadian2010synthetic, stopher2007household}. In such cases, practitioners may need to apply an existing model to a new geographic or survey context before sufficient target-domain data are available for recalibration \citep{sikder2013spatial}. The external dataset experiment examines how well each modeling approach transfers when no target-domain adaptation is performed.

}

\begin{table}[!ht]
\caption{Performance of baseline models and LLMs under different RAG strategies in two external datasets. The best-performing value is highlighted in \textbf{bold}, and the second-best value is \underline{underlined}.}
\label{tab:generalization}
\renewcommand{\arraystretch}{1.4} 
\setlength{\tabcolsep}{5pt} 
\centering
\begin{adjustbox}{width=\textwidth,center}
\begin{tabular}{p{2cm}p{5cm}|cccc|cccc}
\Xhline{1.2pt} 
                         &                                       & \multicolumn{4}{c|}{\textbf{2023   PSRC - Tacoma}}                           & \multicolumn{4}{c}{\textbf{2022   NHTS}}                                     \\ \cline{3-10} 
\textbf{Model}           & \textbf{Method}                       & \textbf{Accuracy} & \textbf{F1 Score} & \textbf{Recall} & \textbf{Precision} & \textbf{Accuracy} & \textbf{F1 Score} & \textbf{Recall} & \textbf{Precision} \\ \hline
MNL                      & -                                     & 0.236             & 0.140             & 0.236           & 0.817              & 0.100             & 0.049             & 0.100           & 0.034              \\
RF                       & -                                     & 0.408             & 0.432             & 0.408           & 0.755              & 0.364             & 0.454             & 0.364           & 0.827              \\
MLP                      & -                                     & 0.345             & 0.359             & 0.344           & 0.790              & 0.322             & 0.420             & 0.322           & 0.851              \\ \hline
\multirow{5}{*}{GPT-4o}  & Zero-shot                             & 0.807             & 0.810             & 0.807           & 0.848              & 0.858             & 0.861             & 0.858           & 0.870              \\
                         & Basic RAG                             & \underline{0.844}       & 0.831             & \underline{0.844}     & 0.846              & \textbf{0.896}    & \textbf{0.882}    & \textbf{0.896}  & \underline{0.872}        \\
                         & RAG + Balanced Retrieval              & 0.833             & 0.826             & 0.833           & 0.832              & 0.891             & \underline{0.881}       & 0.891           & 0.871              \\
                         & RAG + Re-ranking                      & 0.831             & 0.824             & 0.831           & 0.832              & 0.882             & 0.874             & 0.882           & 0.866              \\
                         & RAG + Balanced Retrieval + Re-ranking & 0.838             & 0.834             & 0.838           & 0.838              & 0.877             & 0.871             & 0.877           & 0.866              \\ \hline
\multirow{5}{*}{o4-mini} & Zero-shot                             & 0.831             & 0.829             & 0.831           & \underline{0.854}        & 0.882             & 0.878             & 0.882           & \textbf{0.875}     \\
                         & Basic RAG                             & 0.833             & 0.826             & 0.833           & 0.836              & 0.893             & 0.876             & 0.893           & 0.867              \\
                         & RAG + Balanced Retrieval              & 0.815             & 0.807             & 0.815           & 0.808              & 0.873             & 0.862             & 0.873           & 0.853              \\
                         & RAG + Re-ranking                      & 0.831             & 0.823             & 0.831           & 0.827              & 0.879             & 0.870             & 0.879           & 0.862              \\
                         & RAG + Balanced Retrieval + Re-ranking & 0.805             & 0.803             & 0.805           & 0.801              & 0.849             & 0.842             & 0.849           & 0.838              \\ \hline
\multirow{5}{*}{o3}      & Zero-shot                             & \textbf{0.870}    & \textbf{0.859}    & \textbf{0.870}  & \textbf{0.863}     & 0.880             & 0.864             & 0.880           & 0.854              \\
                         & Basic RAG                             & 0.837             & 0.826             & 0.837           & 0.842              & \underline{0.895}       & 0.876             & \underline{0.895}     & 0.867              \\
                         & RAG + Balanced Retrieval              & 0.828             & 0.821             & 0.828           & 0.827              & 0.882             & 0.872             & 0.882           & 0.863              \\
                         & RAG + Re-ranking                      & 0.828             & 0.820             & 0.828           & 0.830              & 0.875             & 0.866             & 0.875           & 0.859              \\
                         & RAG + Balanced Retrieval + Re-ranking & \underline{0.844}       & \underline{0.838}       & \underline{0.844}     & 0.834              & 0.856             & 0.847             & 0.856           & 0.840              \\ \Xhline{1.2pt}
\end{tabular}
\end{adjustbox}
\end{table}

On the 2023 PSRC-Tacoma subset, all three LLM-based models (GPT-4o, o4-mini, o3) demonstrated substantial improvements over traditional baselines (MNL, RF, MLP), achieving accuracies well above 0.80 compared to below 0.41 for conventional models. Among the LLMs, the o3 model achieved the strongest zero-shot performance, with an accuracy of 0.870 and F1 score of 0.859, outperforming all RAG-augmented configurations. This result is consistent with the earlier finding that o3 possesses strong intrinsic {\YX zero-shot transfer} capabilities.
For GPT-4o, retrieval augmentation proved more beneficial. Accuracy improved from 0.807 in zero-shot to 0.844 with Basic RAG and further to 0.838 with the combination of Balanced Retrieval and Re-ranking, indicating that while zero-shot transfer is relatively strong, the inclusion of targeted retrieval helps the model better exploit contextual information when the external dataset is close in nature to the training distribution. 
By contrast, o4-mini exhibited more modest {\YX no-retraining transferability}, with zero-shot accuracy at 0.831 and mixed gains from retrieval augmentation, peaking at 0.833 with Basic RAG. 
Overall, the Tacoma results highlight that both strong zero-shot reasoning and well-calibrated retrieval strategies enable successful {\YX no-retraining transfer}.

For the 2022 NHTS subset, traditional baselines again performed poorly, with accuracies below 0.37, indicating the difficulty of transferring statistical and machine learning approaches without retraining. In contrast, LLMs {\YX performed} substantially better. Here, RAG played a more critical role than in 2023 PSRC-Tacoma subset. For GPT-4o, accuracy improved from 0.858 in zero-shot to 0.896 with Basic RAG, the best performance across all models and strategies on this dataset. 
Similarly, o3 achieved its peak NHTS accuracy (0.895) with Basic RAG. Interestingly, Balanced Retrieval and re-ranking provided only marginal or even negative gains in this setting, suggesting that over-constraining the retrieval pool may reduce adaptability under greater distribution shifts.
The o4-mini model again showed intermediate performance, with accuracy ranging from 0.882 in zero-shot to 0.893 with Basic RAG. While its absolute performance lagged behind GPT-4o and o3, it still outperformed conventional baselines by a wide margin.
Notably, across all LLMs, the best-performing strategies on NHTS relied on broader retrieval (i.e., Basic RAG) rather than more refined approaches, suggesting that under a significant distribution shift, maximizing coverage of candidate examples is more valuable than enforcing balance or precision.

\section{Discussion}
\label{s:6}

The results of this study offer significant insights into the application of LLMs and RAG for travel mode choice prediction, highlighting a complex interplay between LLM architecture and retrieval strategies. A key takeaway is that retrieval augmentation is not universally beneficial; its effectiveness depends heavily on the reasoning capability of the base model. The performance of the o3 model serves as an illustration of this principle. Its accuracy decreased with the introduction of basic and balanced RAG strategies, suggesting that for a model with advanced reasoning, low-quality or irrelevant retrieved information acts as noise, hindering rather than helping its predictive performance. This phenomenon mirrors a human expert's cognitive process: an expert's judgment is refined by high-quality, relevant evidence, but can be confused by noisy or misleading information \citep{kahneman2021noise}. It was only when a highly sophisticated retrieval pipeline (e.g., RAG with a cross-encoder for re-ranking) was employed that the o3 model's performance improved beyond its zero-shot baseline. This demonstrates that for high-reasoning models, the quality of the retrieved context is paramount. In contrast, models like GPT-4o and o4-mini, which started from lower zero-shot baseline, benefited from even the simpler RAG strategies, as the retrieved examples provided a contextual grounding that their inherent knowledge base may have lacked.

The task of predicting travel mode choice is more than a simple classification problem. It is a simulation of complex human decision-making, requiring the consideration of numerous factors such as cost, time, convenience, and personal preferences. The superior performance of the LLM-based approaches demonstrates the importance of a model's ability to engage in sophisticated reasoning. The exceptional zero-shot accuracy of the o3 model is particularly telling. It demonstrates that a model with strong reasoning capabilities can effectively capture the complex trade-offs in travel decisions without observing examples from the dataset. This suggests that the model is not merely pattern-matching based on the input features but is applying a more generalized, abstract understanding of human behavior to the specific context of the trip. This capacity for contextual reasoning is a significant advantage over traditional statistical models, which are constrained by rigid assumptions, and machine learning models that excel at pattern recognition but lack genuine comprehension.

Based on the experimental results, several best practices emerge for researchers and practitioners looking to leverage these technologies. First, retrieval strategies should be aligned with model capability. The choice of a RAG strategy should not be made independently of the choice of the LLM. For highly advanced, powerful reasoning models like o3, investing in a high-precision retrieval and re-ranking pipeline is crucial. For less advanced or more cost-effective models, a simpler RAG implementation can still provide substantial performance gains over a zero-shot approach. Second, establishing a strong zero-shot baseline is critical. Before implementing a complex RAG system, the zero-shot performance of the candidate LLM should be evaluated. As seen with the o3 model, a high-performing base model may set a high bar for any augmentation strategy to surpass. This baseline provides a critical reference point for assessing the true value added by the retrieval pipeline. Third, retrieval design should account for the distributional characteristics of the target dataset. When the external data closely resembles the knowledge base distribution, advanced strategies such as balanced retrieval and cross-encoder re-ranking can refine predictions. However, under greater distributional shifts, broader retrieval methods like Basic RAG may prove more effective.

This study demonstrates that thoughtfully integrating LLMs and RAG can significantly advance travel behavior modeling. By grounding the reasoning capabilities of LLMs in empirical data through RAG, we can create predictive tools that are both more accurate and transparent. Furthermore, the flexibility of the RAG framework allows for the dynamic incorporation of new data, making it possible to create models that can adapt to changing travel patterns, new transportation options, or unforeseen events in ways that statically trained models cannot.

\section{Conclusion}
\label{s:7}

This study evaluated various models for the task of travel mode choice prediction, comparing traditional baselines and LLMs augmented with a range of RAG strategies. By systematically testing models from the highly capable GPT-4o and o3 to the more compact o4-mini model under zero-shot, basic RAG, and advanced RAG conditions, this research sought to identify the most effective approaches for simulating the complex, multi-faceted decisions in human travel mode choice.

The results indicate that the GPT-4o model achieved the highest overall accuracy of 80.8\% using a RAG with balanced retrieval and re-ranking strategy. We also find that retrieval augmentation is not a universal benefit. For a model with high intrinsic reasoning like o3, basic and moderately improved RAG pipelines introduced noise and degraded performance. Only when paired with a high-precision re-ranking approach did o3 surpass its already strong zero-shot baseline. This finding indicates that the effectiveness of retrieval augmentation is contingent upon the reasoning capacity of the underlying LLM: while less advanced models can benefit substantially from simpler retrieval strategies, advanced models require only the most precise and contextually relevant evidence to realize performance gains. In summary, the optimal configuration of LLMs and RAG strategies is not uniform but rather depends on both model architecture and task objectives. For researchers and practitioners, this highlights the importance of aligning retrieval design with model capability and application context in order to fully unlock the potential of RAG-augmented LLMs for travel behavior modeling.

While this study provides valuable insights, it is subject to certain limitations. First, while we infer reasoning capabilities from performance patterns, this research did not directly analyze the models' internal Chain-of-Thought processes. Second, the RAG pipelines were limited to a specific set of components, and other variations in retrieval, chunking, or re-ranking could yield different results. Future research should therefore aim to validate these findings across more diverse datasets, employ methods that analyze the step-by-step reasoning of these models, and explore a broader array of RAG architectures and their real-world deployment implications.



\section*{Declaration of Generative AI and AI-assisted Technologies in the Writing Process}
During the preparation of this work the authors used ChatGPT 5 in order to improve readability and language. After using this tool/service, the authors reviewed and edited the content as needed and take full responsibility for the content of the publication.

\section*{Declaration of Competing Interest}
None.

\appendix
\section{Descriptive statistics of variables for generalization ability evaluation}

\begin{table}[H]
\caption{Descriptive statistics of variables for generalization ability evaluation}
\label{tab:var_stat_gen}
\renewcommand{\arraystretch}{1.2}
\setlength{\tabcolsep}{5pt} 
\begin{adjustbox}{width=\textwidth,center}
\begin{tabular}{llllll}
\Xhline{1.2pt}
                        &                                 & \multicolumn{2}{c}{\textbf{2023   PSRC - Tacoma}} & \multicolumn{2}{c}{\textbf{2022   NHTS}} \\ \cline{3-6} 
Variable                & Value                           & Frequency/Mean      & Percentage/SD      & Frequency/Mean  & Percentage/SD \\ \hline
\multicolumn{6}{l}{\textbf{Outcome Variable}}                                                                                                   \\
Trip Mode               & Drive                           & 409                 & 71.88\%            & 485             & 85.24\%       \\
                        & Walk                            & 127                 & 22.32\%            & 61              & 10.72\%       \\
                        & Transit                         & 23                  & 4.04\%             & 13              & 2.28\%        \\
                        & Bike/Mircomobility              & 10                  & 1.76\%             & 10              & 1.76\%        \\ \hline
\multicolumn{6}{l}{\textbf{Traveler   Characteristics}}                                                                                         \\
Age                     & 18-24                           & 56                  & 9.84\%             & 19              & 3.34\%        \\
                        & 25-34                           & 158                 & 27.77\%            & 80              & 14.06\%       \\
                        & 35-44                           & 118                 & 20.74\%            & 118             & 20.74\%       \\
                        & 45-54                           & 158                 & 27.77\%            & 97              & 17.05\%       \\
                        & 55-64                           & 23                  & 4.04\%             & 112             & 19.68\%       \\
                        & 65-74                           & 35                  & 6.15\%             & 102             & 17.93\%       \\
                        & 75 or older                     & 21                  & 3.69\%             & 41              & 7.21\%        \\
Gender                  & Male                            & 279                 & 49.03\%            & 313             & 55.01\%       \\
                        & Female                          & 265                 & 46.57\%            & 256             & 44.99\%       \\
                        & Non-binary                      & 25                  & 4.39\%             & 0               & 0\%           \\
Education   Level       & Less than high school           & 2                   & 0.35\%             & 6               & 1.05\%        \\
                        & High school                     & 66                  & 11.60\%            & 34              & 5.98\%        \\
                        & Some college                    & 104                 & 18.28\%            & 71              & 12.48\%       \\
                        & Vocational/technical   training & 20                  & 3.51\%             & 0               & 0\%           \\
                        & Associates degree               & 106                 & 18.63\%            & 63              & 11.07\%       \\
                        & Bachelor degree                 & 169                 & 29.70\%            & 221             & 38.84\%       \\
                        & Graduate degree                 & 102                 & 17.93\%            & 174             & 30.58\%       \\
Household   Income      & Under \$25,000                  & 54                  & 9.49\%             & 18              & 3.16\%        \\
                        & $25,000-$49,999                 & 56                  & 9.84\%             & 62              & 10.90\%       \\
                        & $50,000-$74,999                 & 163                 & 28.65\%            & 84              & 14.76\%       \\
                        & $75,000-$99,999                 & 101                 & 17.75\%            & 98              & 17.22\%       \\
                        & $100,000-$199,999               & 129                 & 22.67\%            & 204             & 35.85\%       \\
                        & \$200,000 or more               & 66                  & 11.60\%            & 103             & 18.10\%       \\
Vehicles in   Household &                                 & 1.51                & 0.95               & 1.90            & 0.91          \\ \hline
\multicolumn{6}{l}{\textbf{Trip Characteristics}}                                                                                               \\
Trip Distance   (mile)  &                                 & 2.26                & 2.02               & 6.29            & 6.44          \\
Start Time              &                                 & 13.65               & 4.34               & 13.31           & 4.31          \\
Trip Purpose            & Home                            & 202                 & 35.50\%            & 231             & 40.60\%       \\
                        & Work                            & 65                  & 11.42\%            & 64              & 11.25\%       \\
                        & Social/Recreation               & 97                  & 17.05\%            & 75              & 13.18\%       \\
                        & Shopping                        & 86                  & 15.11\%            & 98              & 17.22\%       \\
                        & Meal                            & 41                  & 7.21\%             & 40              & 7.03\%        \\
                        & Business/Errand                 & 24                  & 4.22\%             & 17              & 2.99\%        \\
                        & Escort                          & 43                  & 7.56\%             & 31              & 5.45\%        \\
                        & Overnight                       & 6                   & 1.05\%             & 0               & 0\%           \\
                        & School                          & 5                   & 0.88\%             & 13              & 2.28\%        \\
                        & Change mode                     & 0                   & 0\%                & 0               & 0\%           \\ \Xhline{1.2pt}
\end{tabular}
\end{adjustbox}
\end{table}

\bibliographystyle{elsarticle-harv}
\biboptions{semicolon,round,sort,authoryear}
\bibliography{cas-refs}


\end{document}